\documentclass{article} 
\usepackage{iclr2025_conference,times}

\usepackage{amsmath,amsfonts,bm}

\newcommand{\rsm}{64$\times$64}
\newcommand{\net}{N}
\newcommand{\prm}{\theta}
\newcommand{\x}{x}
\newcommand{\y}{y}
\newcommand{\xs}{x_0}
\newcommand{\xt}{x_1}
\newcommand{\ms}{\mu_0}
\newcommand{\mt}{\mu_1}

\newcommand{\ps}{p_0}
\newcommand{\pt}{q}
\newcommand{\vel}{v}
\newcommand{\bs}{{n}}

\newcommand{\batch}{B_1}
\newcommand{\sdim}{d}
\newcommand{\dotp}[2]{\langle #1, #2 \rangle}
\newcommand{\Dotp}[2]{\bigl< #1, #2 \bigr>}
\newcommand{\Ex}{\mathbb{E}}
\newcommand{\Vr}{\mathbb{V}}
\newcommand{\Rn}{\mathbb{R}}
\newcommand{\nrml}{\mathcal{N}}
\newcommand{\dis}{D}
\newcommand{\smp}{\text{Sampler}}

\def\eqref#1{equation~\ref{#1}}

\def\1{\bm{1}}

\DeclareMathAlphabet{\mathsfit}{\encodingdefault}{\sfdefault}{m}{sl}
\SetMathAlphabet{\mathsfit}{bold}{\encodingdefault}{\sfdefault}{bx}{n}

\usepackage{hyperref}
\usepackage{url}
\usepackage{graphicx}
\usepackage{booktabs}
\usepackage{rotating}
\usepackage{booktabs}

\usepackage{booktabs}
\usepackage{multirow}
\usepackage{wrapfig}

\usepackage{subcaption}

\title{Generative Lines Matching Models}

\author{Ori Matityahu and Raanan Fattal \\
School of Computer Science and Engineering\\
The Hebrew University of Jerusalem, Israel\\
\texttt{\{ori.matityahu,raanan.fattal\}@mai.huji.ac.il} \\
}

\iclrfinalcopy 
\begin{document}

\maketitle

\begin{abstract}

In this paper we identify the source of a singularity in the training loss of key denoising models, that causes the denoiser's predictions to collapse towards the mean of the source or target distributions. This degeneracy creates false basins of attraction, distorting the denoising trajectories and ultimately increasing the number of steps required to sample these models.

We circumvent this artifact by leveraging the deterministic ODE-based samplers, offered by certain denoising diffusion and score-matching models, which establish a well-defined change-of-variables between the source and target distributions. Given this correspondence, we propose a new probability flow model, the \emph{Lines Matching Model} (LMM), which matches globally straight lines interpolating the two distributions. We demonstrate that the flow fields produced by the LMM exhibit notable temporal consistency, resulting in trajectories with excellent straightness scores.

Beyond its sampling efficiency, the LMM formulation allows us to enhance the fidelity of the generated samples by integrating domain-specific reconstruction and adversarial losses, and by optimizing its training for the sampling procedure used. Overall, the LMM achieves state-of-the-art FID scores with minimal NFEs on established benchmark datasets: 1.57/1.39 (NFE=1/2) on CIFAR-10, 1.47/1.17 on ImageNet \rsm, and 2.68/1.54 on AFHQ \rsm.

Finally, we provide a theoretical analysis showing that the use of optimal transport to relate the two distributions suffers from a curse of dimensionality, where the pairing set size (mini-batch) must scale exponentially with the signal dimension.
 
\end{abstract}

\section{Introduction} 
\label{sec:intro}

Diffusion models are the core engine behind many recent state-of-the-art generative models across various domains, e.g., image generation~\citep{Song2021SMLD,Ho2020DDPM,Dhariwal2021beatgan,rombach2022high}, text-to-image generation~\citep{Nichol2022text2img,Ramesh2022text2img,Saharia2024}, audio synthesis~\citep{kong2021diffwave,Kim2021GuidedTTSAD, chen2020wavegrad, popov2021gradtts}, and video generation~\citep{ho2022video,singer2023makeavideo,Liu2024sora}

This gain in popularity of the underlying denoising diffusion~\citep{sohl2015thermo,Ho2020DDPM} and score-matching~\citep{song2019sliced,song2020improved,Song2021NCSN} models over GANs~\citep{Goodfellow2014GAN} is often attributed to their improved distribution reproduction~\citep{Dhariwal2021beatgan}, and immunity to various optimization hurdles that plague GAN training (mode collapse and forgetting~\citep{Hoang2020modecollapse}). Nevertheless, unlike the single-step sampling of GAN and VAE~\citep{KingmaW2013VAE} models, the noise removal process follows non-trivial probability flow trajectories, requiring fine quadrature steps and resulting in non-negligible computational effort during inference. This ranges between hundreds of sampling steps in early methods~\citep{Ho2020DDPM} and tens in more recent ones~\citep{Tero2022EDM}.

Distilling pre-trained denoising models allows reducing this Number of Function Evaluations (NFEs) during sampling. This approach can be carried out in different ways; learning the entire sampling procedure~\citep{Luhman2021knowdistl}, or reducing its number of steps progressively~\citep{Salimans2022PD}. More recently the denoising trajectories are learned either by ensuring a consistency along successive steps~\citep{Song23CM}, or along arbitrary segments~\citep{Kim2024CTM}. These methods offer a significant speedup over their teacher models, nevertheless, they also inherit inefficiencies inherent to the trajectories that they replicate.

As an alternative, the probability flow matching techniques in~\citep{Lipman2023flow,Eric2023flow} incorporate Optimal Transport (OT) considerations in order to produce more constant flow trajectories, requiring fewer sampling steps. Additional improvement in straightness is achieved by an iterative rectification scheme in~\citep{Liu2023RectFlow,liu2024instaflow}, as well as by replacing the random pairing between the source and data examples with an OT pairing~\citep{Pooladian2023mini,Tong2024mini}. While improving upon traditional denoising losses, the flow fields produced by these approaches still contain false attraction basins, causing the trajectories to curve.

In this paper, we show that the ambiguous pairing between latent source noise and target data samples leads to an ill-posed regression problem, compromising the performance of key denoising models, including denoising diffusion, score- and flow-matching. At low signal-to-noise ratios, this indeterminacy in the denoising loss becomes worse and causes the denoiser's predictions to collapse toward the mean of either the source or target distributions. This creates false basins of attraction that curve and distort the denoising trajectories, ultimately increasing the number of steps needed for accurate sampling.

We avoid this singularity by leveraging the fact that certain denoising diffusion~\citep{Song2021implicit} and score-matching~\citep{Song2021NCSN,Tero2022EDM} models construct \emph{deterministic} ODE-based flows that give rise to a well-defined change-of-variable between the source and target distributions. Unlike existing approaches that distill the underlying inefficient probability flow trajectories, we only leverage the pairing induced between the distributions. Given this correspondence, we construct a new probability flow model, the \emph{Lines Matching Model} (LMM), which matches \emph{globally straight} lines interpolating between the distributions. As demonstrated in Figure~\ref{fig:traj}, the flow fields produced by the LMM display notable temporal consistency, resulting in trajectories with excellent straightness scores.

Beyond its sampling efficiency, and unlike other flow matching formulations, the LMM's training loss allows us to further improve the fidelity of its generated samples by incorporating domain-specific reconstruction
and adversarial losses, as well as optimizing its training for the sampling procedure used. Overall, the LMM achieves state-of-the-art Fréchet Inception Distance (FID) scores using a minimal NFEs on established benchmarks, specifically, 1.57/1.39 (NFE=1/2) for CIFAR-10, 1.47/1.17 for ImageNet \rsm, and 2.8/1.61 for AFHQ \rsm.

In addition, we make a theoretical contribution showing that while the OT-based pairing in~\citep{Pooladian2023mini, Tong2024mini} is a valid approach for reducing the attraction to the false basins, due to a fundamental course-of-dimensionality, the batch size required scales exponentially as a function of the signal dimension. Given that the latter is fairly high across various domains and the former is typically constrained by memory and compute limitations, the effectiveness of this approach is limited, as demonstrated in Figure~\ref{fig:traj}.

\section{Background}
\label{sec:background}

\setlength{\tabcolsep}{0.4pt}
\renewcommand{\arraystretch}{0.5} 
\newcommand{\ffig}[1]{ \includegraphics[width=0.22\textwidth]{#1}}

    \begin{table}
    \vspace{-0.3in}
        \centering
        \begin{tabular}{ccc}
         source dist. &   target dist. & \\
         
             \vspace{0.1in}
             \ffig{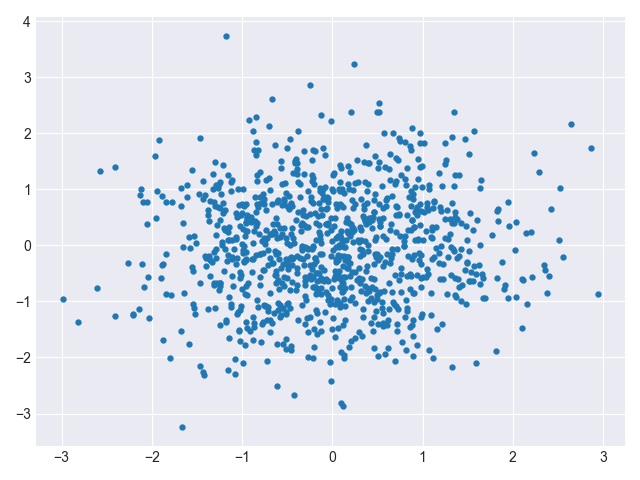} &   \ffig{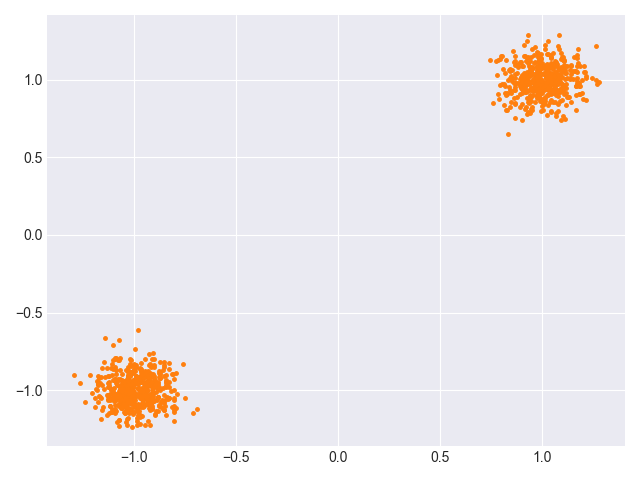} & \raisebox{+3.5em}{ \scriptsize
             \begin{tabular}{lr} 
\toprule
model & straightness \\
\midrule
EDM & 0.0397 \\
OT-CFM & 0.0417 \\
BOT-CFM & 0.030 \\
1/2-RF & 0.043/0.00135 \\
LMM & 0.00183 \\
\bottomrule
\end{tabular} \normalsize }  \\
              initial flow &  mid-step  flow &  sampling trajectories \\
              
              \raisebox{+2.2em}{
                \begin{sideways}
                {EDM}
                \end{sideways}} 
             \ffig{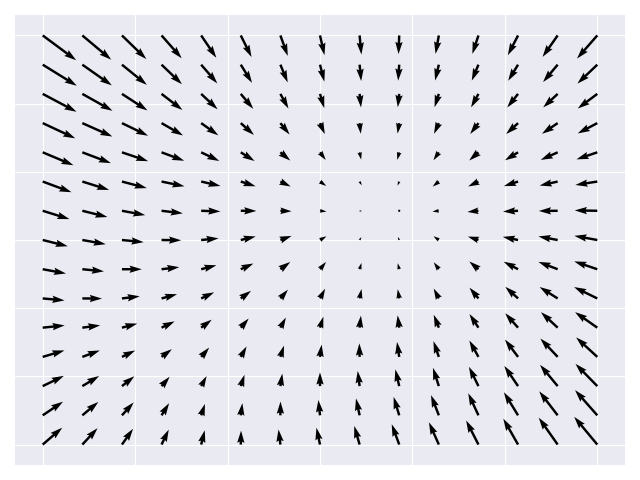} &  \ffig{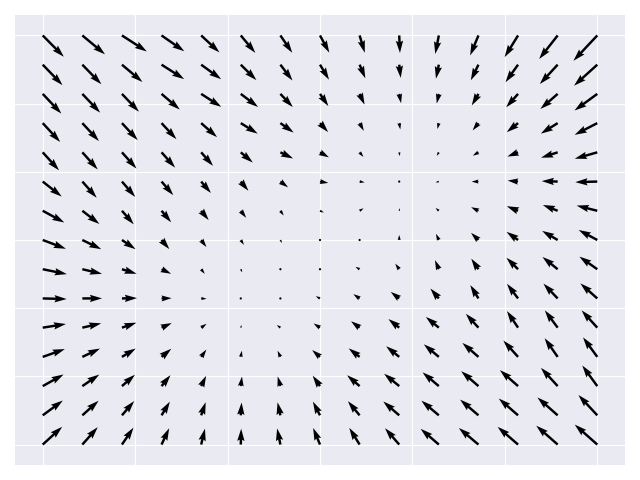} & \ffig{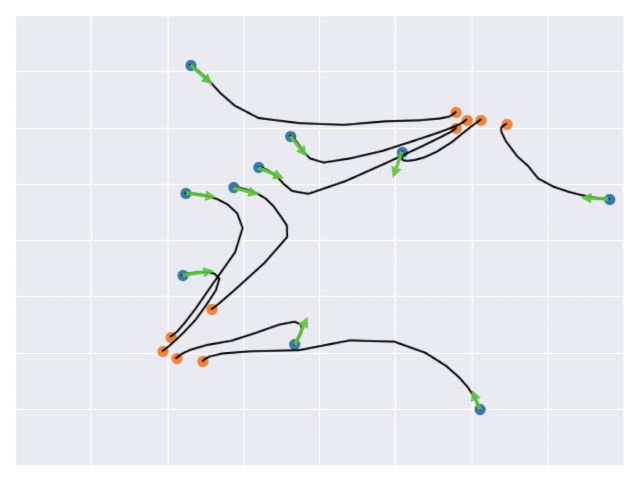}   \\

              \raisebox{+1.5em}{
                \begin{sideways}
                {OT-CFM}
                \end{sideways}} 
            \ffig{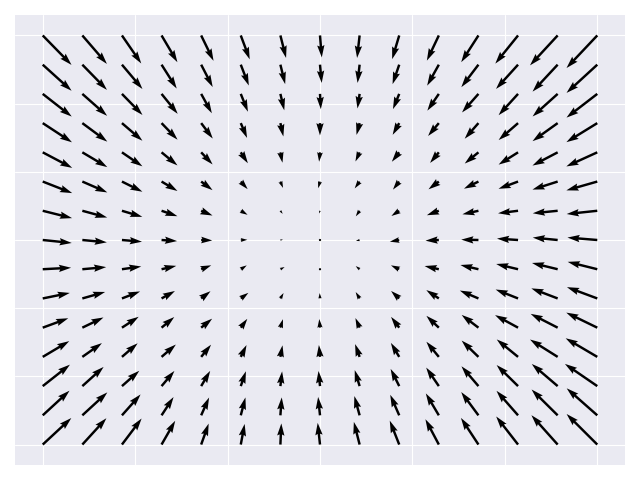} &  \ffig{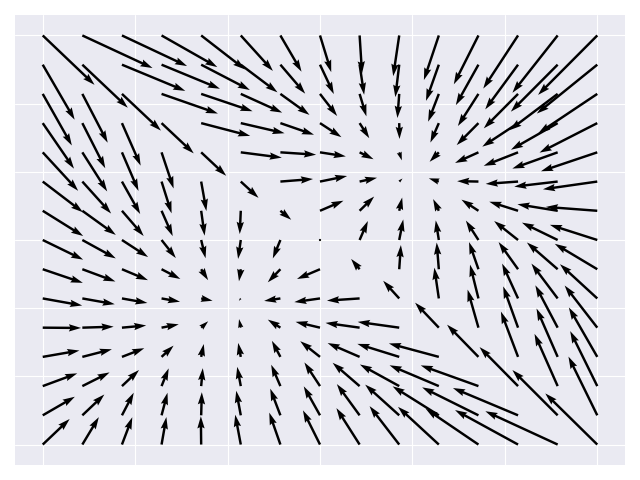} & \ffig{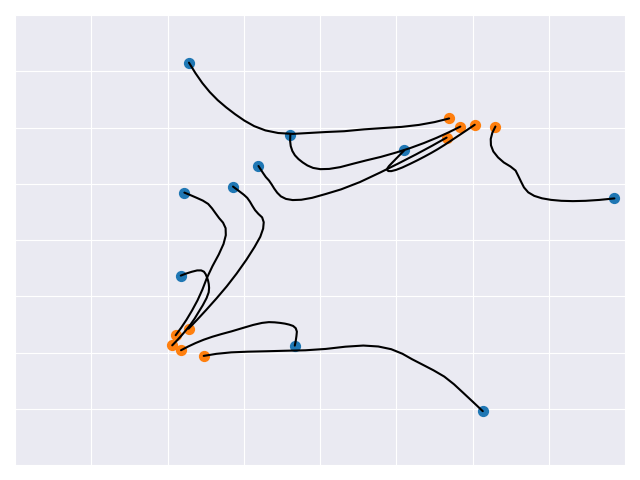}   \\
             
             \raisebox{+1em}{
                \begin{sideways}
                {BOT-CFM}
                \end{sideways}} 
             \ffig{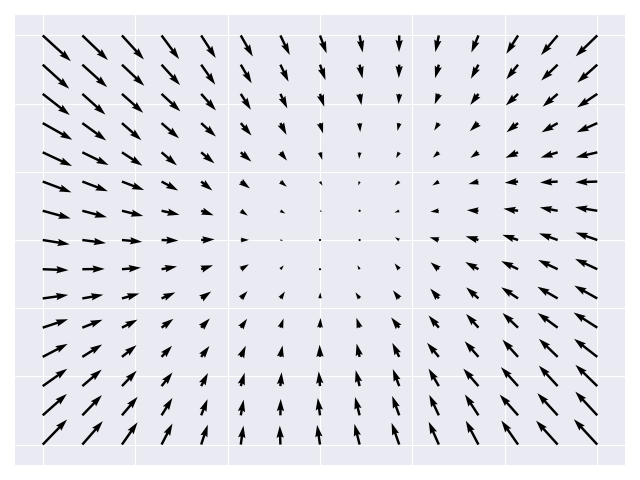} &  \ffig{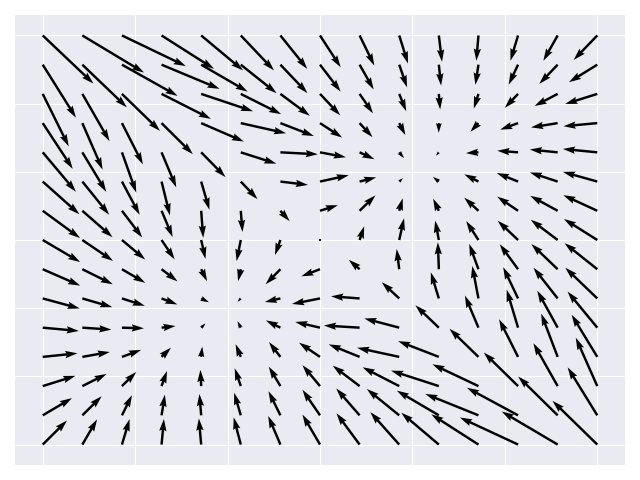} &  \ffig{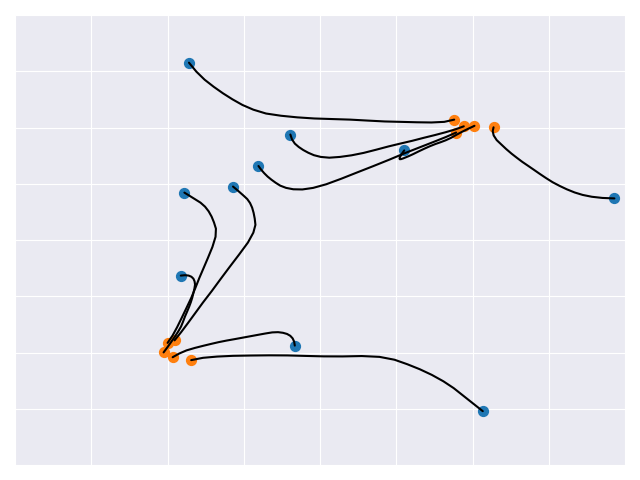}   \\

             \raisebox{+1.2em}{
                \begin{sideways}
                {Rect-Flow}
                \end{sideways}} 
             \ffig{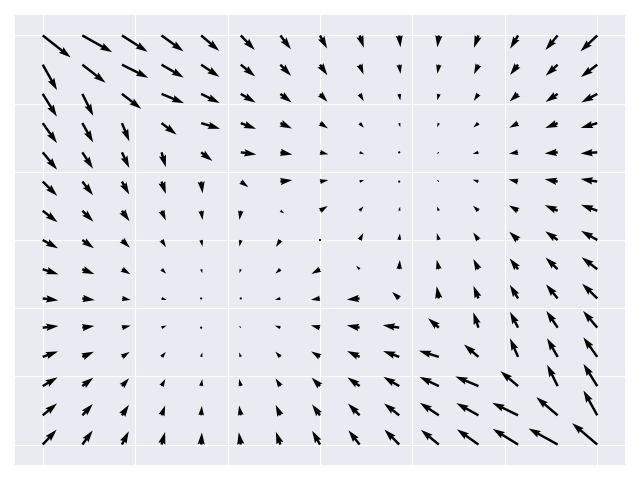} &  \ffig{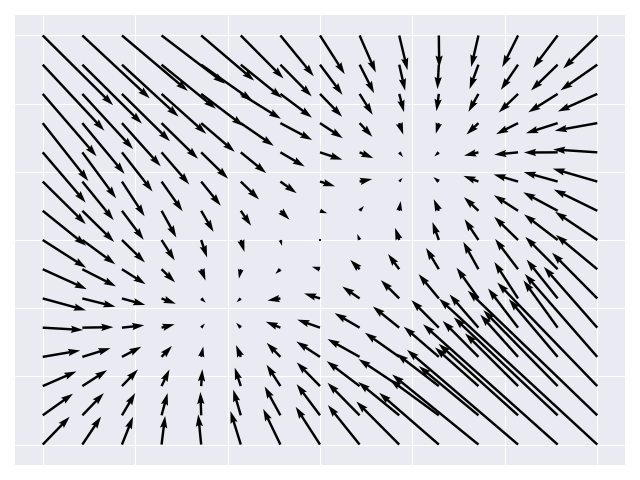} &  \ffig{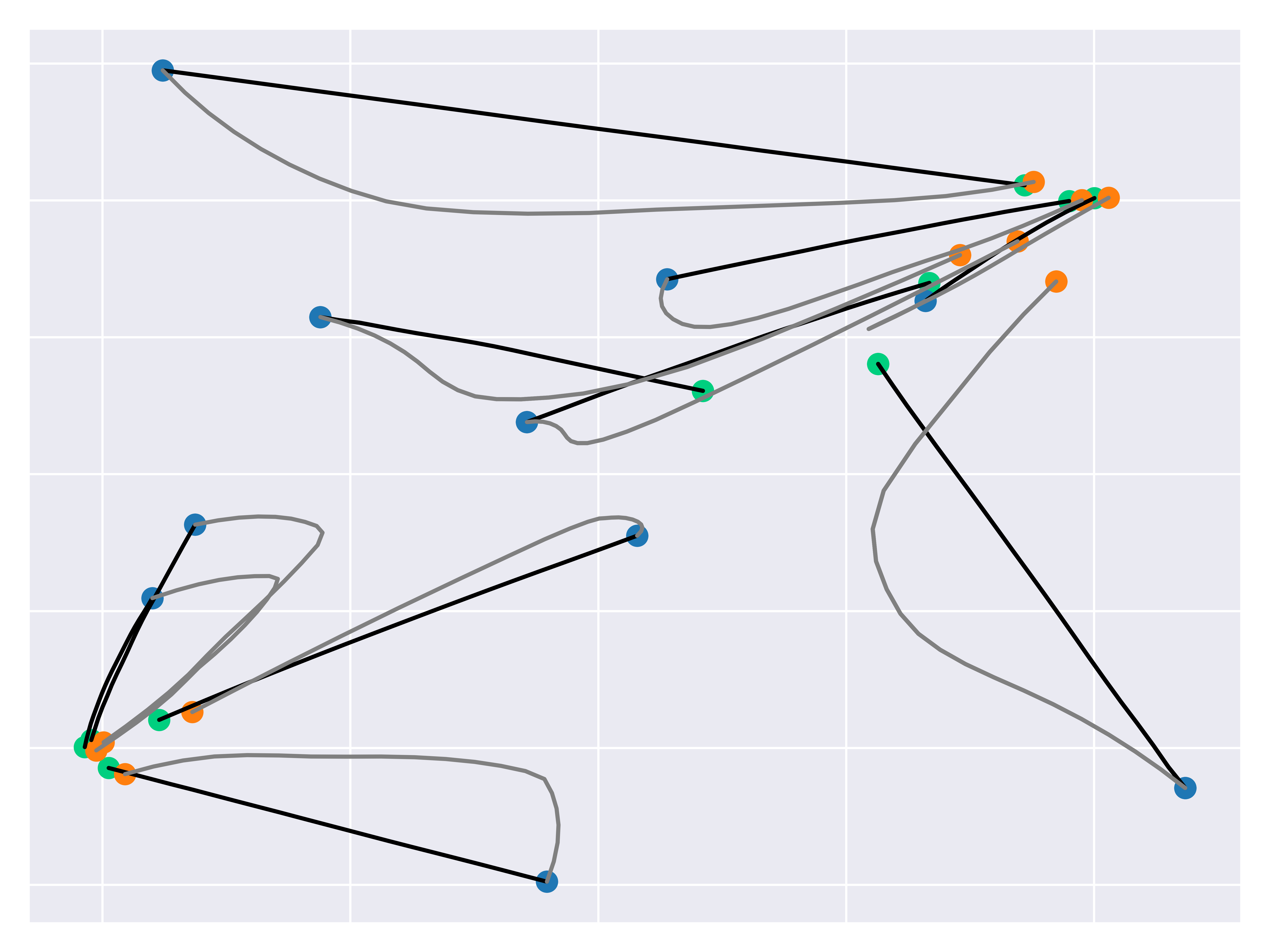}   \\
             
              \raisebox{+2em}{
                \begin{sideways}
                {LMM}
                \end{sideways}} 
              \ffig{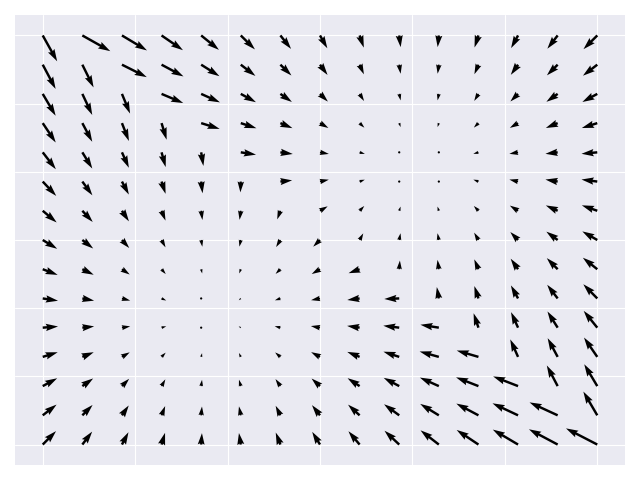}  &  \ffig{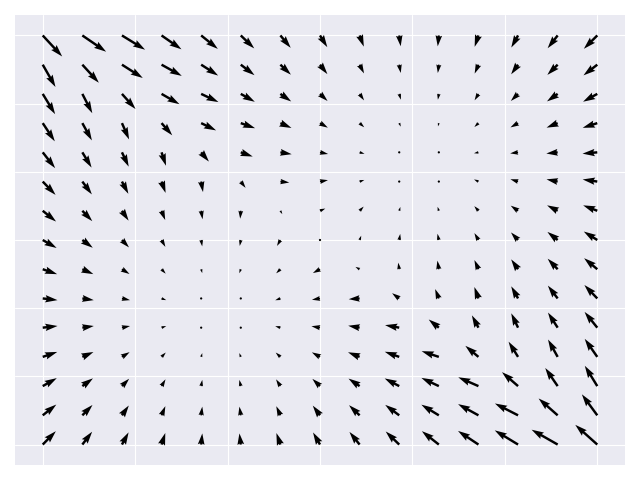} & \ffig{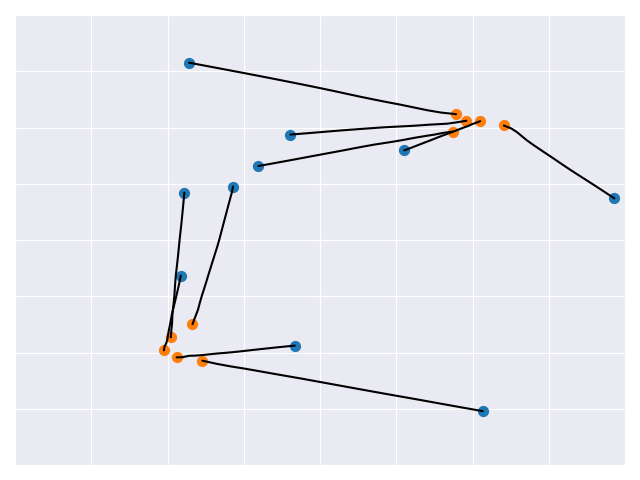}  \\
            
        \end{tabular}
        \caption{Flow fields and sampling trajectories of different models. Top row shows the source and target distributions along their first two dimensions. The source (blue dots) is a normal distribution in 128 space dimension. The target (orange dots) is a mixture of two Gaussians located in $(-1,-1,\vec{0})$ and $(1,1,\vec{0})$ with STDs of $(0,1,0.1,\vec{1})$, i.e., two separate Gaussians in the first two dimensions shown in the figure, and a normal Gaussian in the remaining 126 dimensions. The following three rows show results of the optimized DSM approach of EDM~\citep{Tero2022EDM}, the OT-CFM~\citep{Lipman2023flow} and its mini-batch optimized BOT-CFM~\citep{Pooladian2023mini} which all appear to produce curved trajectories, with an improvement observed in the BOT-CFM when pairing batches of size 256. Nearly identical results are obtained using a batch size of 128, differing in straightness by only $0.0017$. The trajectories of the 1-Rect-Flow~\citep{Liu2023RectFlow}, shown in gray in the next row, also appear curved. The 2-Rect-Flow trajectories (black) are considerably straighter than any of the above. However, a discrepancy between these two iterations can be seen in their (target) endpoints (orange and cyan dots). This may indicate a drift from the original distribution $\pt$. Our LMM produces straight curves and flow Fields which are close to being constant in time. Note that excluding 2-Rect-Flow and LMM, the initial flow fields of all the methods show a clear basin of attraction at $(0,0,\vec{0})$ responsible for an undesired drift at the beginning of the trajectories towards this point. This effect is illustrated in the EDM, where the green arrows represent the tangent vectors to the curves at their initial step. Top-right table reports the average trajectory straightness score, $\int_0^1 \| \dot{\x}(t)-(\xt-\xs) \|dt $, of each method where both 2-Rect-Flow and LMM standout.}
        \label{fig:traj}
    \end{table}

We begin by reviewing several key denoising-based generative models, with an attempt to bring them to a common form in order to highlight the source of a sampling inefficiency that they share, and we address them in our work. The Denoising Diffusion Probability Models (DDPM)~\citep{sohl2015thermo,Ho2020DDPM}, as well as Denoising Score Matching (DSM) approaches, specifically the Noise Conditional Score Network (NCSN)~\citep{Song2021NCSN} use the following form of denoising loss,
\begin{equation}
\label{eq:denoising}
\text{argmin}_\prm \Ex_{t,q(\xt),p(\x|\xt,t)} \Big[  \Vert \net_\prm(\x,s_t) - \nabla_{\x}\log p(\x|\xt,t) \Vert^2 \Big],
\end{equation}
where $\pt(\xt)$ is the target data distribution which we are given empirically. In case of DDPM, $p(\x|\xt,t) = \nrml (\sqrt{\alpha_t}\xt,(1-\alpha_t) I) $ and $s_t = t $, where $1\leq t \leq N$ is a noise scheduling index weighted by probabilities $\propto (1-\alpha_t)$, and $\alpha_t=\prod_{i=1}^t(1-\beta_i)$ and $0<\beta_i<1$ are a pre-defined sequence of noise scales\footnote{The $\alpha_t$ defined here correspond to the $\bar{\alpha}_t$ in the derivation of~\cite{Ho2020DDPM}.}. In this framework the network $\net_\prm$ models the mean of the reverse Gaussian kernels by $p(\x^{t-1} | \x^{t}) = \mathcal{N}((\x^{t} +\beta_t\net_\prm(\x^t,t))/\sqrt{1-\beta_t},\beta_tI)$, which are designed to start their operation from a source distribution, $x^N \sim \ps = \mathcal{N}(0,I)$. In the NCSN, $p(\x|\xt,t) = \nrml (\xt,\sigma^2_tI) $ and $s_t=\sigma_t $, where $\{\sigma_t\}_{t=1}^N$ are positive noise scales, weighted $\propto \sigma_t^2$. In this approach, the network $\net_\prm$ models the score field of noised data densities $p(\x,\sigma_t) = \pt * \mathcal{N}(0,\sigma_t^2 I) $, which is used for gradually denoising samples, starting from $x^N \sim \ps = \mathcal{N}(0,\sigma_N^2 I)$, where $\sigma^2_N >> \Vr[\xt]$. Much has been discussed about the close relation between these two approaches~\citep{Pascal2011connection,Song2021SMLD,Tero2022EDM}. This formalism can be further generalized to cover continuous-time Stochastic Diffusion Equations (SDEs), where the DDPM results in a Variance Preserving (VP) process, and the NCSN in a Variance Exploding (VE) process, see~\citep{Song2021SMLD}.

The noised to clean signal regression problem solved in Eq.~\ref{eq:denoising} is known to \emph{underestimate} the true regression~\citep{kendall1973book,Allan2013noisefit}, due to averaging caused by the noise present in $p(\x|\xt,t)$. At the limit of low Signal-to-Noise Ratio (SNR), i.e., high noise level $\sigma_t$ and $\beta_t$ (low $\alpha_t$) at large $t$ in Eq.~\ref{eq:denoising}, where $p(\x|\xt,t) \approx \ps(\x)$, the regression collapses to a constant prediction, specifically $\net(\x,N) \approx \Ex_{\ps}[\xt]=\mu_{\ps}=0 $ in the DDPM, and $\net(\x,\sigma_N)  \approx \Ex_{\pt}[\xt]= \mu_{\pt} $ in the NCSN, as shown in Appendix~\ref{append:reg}. Consequently, rather than moving towards particular instances in the target distribution, the initial sampling steps appear either stationary in the case of DDPM, or gravitate towards $\mu_{\pt}$ in the NCSN, as indicated by the green arrows in Table~\ref{fig:traj}. These instance-independent basins of attraction create inefficient sampling trajectories that lack constancy in speed and direction as also shown in the table. The less constant in speed or direction these trajectories are, the more integration steps are needed to follow them during sampling.

Indeed, rectifying the trajectories towards fixed-speed straight lines, is an important design principle shared by recent flow-based models. The Conditional Flow Matching (CFM) method in~\citep{Lipman2023flow}, constructs a deterministic time-dependent change-of-variable $ \psi(\x,t) $ that gradually maps the source distribution $\ps$ to the target $\pt$. Similarly to the way tractable reverse diffusion kernels are derived in DDPM~\citep{sohl2015thermo}, these maps are constructed by defining simpler conditional maps $\psi_{\xt}(\x,t)= (1-t)\x+t\xt $ that map $\ps$ (a normalized Gaussian), towards a small Gaussian\footnote{To simplify derivation we assume a zero width target Gaussian around each data point, i.e., $\sigma_{\min}=0$ in the formalism of~\citep{Lipman2023flow}} centered around each $\xt$ as a function of $t\in[0,1]$. The network $\net$ is then trained to match an aggregated velocity flow field by marginalizing $\partial \psi_{\xt} / \partial t$ over all the data points $\xt$ by solving,
\begin{equation}
\label{eq:cfm}
  \text{argmin}_\prm \Ex_{t,q(\xt),p(\xs)} \Big[ \Vert \net_\prm\big(t\xt+(1-t)\xs,t \big) - (\xt-\xs) \Vert^2 \Big],
\end{equation}
As in Eq.~\ref{eq:denoising} above, the network is regressed under severe marginalization where the mapping of every $\xs$ to every $\xt$ are averaged together at a non-trivial contribution at low values of $t$. In Appendix~\ref{append:reg} we show that, similarly to the NCSN, this approach also results in $\net_\prm(\x,0) = \mt$ and curved trajectories, shown in Table~\ref{fig:traj}. An alternative derivation in~\citep{Albergo2023stointrp} discusses the option of optimizing the transport of their maps and proposes an initial direction to shorten their path length.


A flow rectification process described in~\citep{Liu2023RectFlow} also matches the flow using Eq.~\ref{eq:cfm}, however it operates iteratively; at each step $k$ it trains $\net^k$ over a different set of source $Z_0^k$ and target $Z_1^k$ examples. The process starts with the \emph{random} pairing used in~\citep{Lipman2023flow}, i.e., $Z_0^1 $ and $Z_1^1$ are independent samples from $\ps $ and $ \pt$ respectively. In the following steps, $Z_0^{k+1}$ and $Z_1^{k+1}$ are produced by generating new samples using $\net^k$ starting from $\ps$ and $\pt$ (by integrating $-\net^k$). This results in a \emph{deterministic} pairing and this process is shown to monotonically increase the straightness of the trajectories in $\net^k$.

As shown in Table~\ref{fig:traj}, the resulting flow trajectories at $k\!=\!1$ share a similar gravitation towards $\mt$ as in the CFM. At $k\!=\!2$ they become significantly more straight and easier to integrate. As $k$ increases errors in the estimated flow field $\net^k$ accumulate and cause $Z_0^k$ and $Z_1^k$ to drift away from $\ps$ and $\pt$ respectively. 2-Rect-Flow ($k\!=\!2$) is said to be found optimal in~\citep{Liu2023RectFlow}.

For completeness, let us note that that deterministic probability flow ODE models were also derived in the contexts of DDPM and DSM. Specifically, the Denoising Diffusion Implicit Model (DDIM)~\citep{Song2021implicit} derives a non-Markovian process, where the inverse kernels map noised samples along deterministic lines with noise-free endpoints. In connection to neural ODEs~\citep{Chen2018CNF}, it is shown in~\citep{Song2021SMLD} that the DSM with Langevin Dynamics (SMLD), which is an SDE, has a deterministic time-reversal process, and following several design improvements this deterministic procedure achieve impressive results in~\citep{Tero2022EDM}. The straightness of the trajectories is not explicitly considered in these works.

\bf{Variance Reduction.}  \normalfont The non-negligible association between \emph{every} pair of samples $\xs\sim\ps$ and $\xt\sim\pt$ when marginalizing the regression losses over an independent distribution $\ps(\xs)\pt(\xt)$, is a common thread shared by all the models mentioned above, which undermines their sampling efficiency. As a remedy, recent works aim to replace this arbitrary pairing with ones that improve sampling.

By linking the flow's transport optimality with the straightness of their trajectories, both~\citep{Pooladian2023mini} and~\citep{Tong2024mini} derive the pairing between $\ps$ and $\pt$ from an Optimal Transport (OT) objective. Due to the cubic complexity of this problem~\citep{Flamary21POT} (or a quadratic approximation~\citep{Altschuler2017sink}) the pairing, or plan $j_i$, is computed within batches of samples $\{\xs^i\}_{i=1}^\bs \sim \ps $ and $\{\xt^i\}_{i=1}^\bs \sim \pt $ of moderate sizes ($\bs=50/256$ in~\citep{Pooladian2023mini}), and Eq.~\ref{eq:cfm} is minimized over these permuted pairs. As shown in Table~\ref{fig:traj} this approach results in flows that less curved than those produced by~\citep{Lipman2023flow}. Indeed, a 30\% to 60\% reduction in sampling cost is reported in~\citep{Pooladian2023mini}.

A well-known manifestation of the curse-of-dimensionality causes the ratio between the farthest and closest points to converge to a constant as the space dimension increases~\citep{Beyer1999NN}. Thus, considerably larger batches are needed in order to find meaningful plans $\pi_{ij}$. In Appendix~\ref{append:OTB} we provide an asymptotic analysis showing an exponential batch size $\bs$ dependency over the space dimension $\sdim$ when solving a rather simpler problem; transporting a unit Gaussian distribution to itself. This finding undermines the prospect of accelerating sampling by increasing the batch size and relying solely on OT pairing. Indeed, in the example shown in Table~\ref{fig:traj}, a negligible difference in trajectory straightness is found between $\bs=128$ and $\bs=256$.

Another strategy to avoid independent pairing described in~\citep{Lee2023VAEtraj} draws $ \xs \sim q(\xs|\xt)$ given $\xt \sim \pt $, where $q$ is a VAE-based encoder that maps points $\xt$ to Gaussians. In order to obtain non-trivial pairing one would seek highly distinctive $q(\xs|\xt)$ for each $\xt$ however, similarly to VAE training, Gaussians of different $\xt$ are trained to match the same $\ps$ in order to be consistent with sampling time.
\section{Lines Matching Models}
\label{sec:method}

As discussed above various diffusion, score and flow matching models achieve a remarkable sampling accuracy in various data domains. This comes at a cost of executing multiple sampling steps at inference time---a notable drawback compared to the single feed-forward execution of a VAE and GAN networks. The inefficiency is rooted in the unfocused association between $\ps$ and $\pt$ produced by the independent example pairing, leading to poorly-resolved denoising, score and flow regression problems in the low SNR regime. Unlike methods that distill these inefficient curved trajectories~\citep{Salimans2022PD,Song23CM,Kim2024CTM}, we only utilize the pairing they induce between the source and target distributions to construct a new probability flow model that matches \emph{globally straight} lines connecting the two distributions.

We derive the Lines Matching Model (LMM) in accordance with the VE probability flow ODE formulation used in~\citep{Tero2022EDM}, by training a neural model $\net_\theta$ to minimize
\begin{equation}
\label{eq:ours}
  \mathcal{L}_\text{lines} = \Ex_{\sigma,\delta(\xt,\psi^*(\xs)),\ps(\xs)} \Big[ \Vert \net_\prm\big(\xt+\sigma \xs, \sigma \big) - \xt \Vert_{\mathcal{P}} \Big],
\end{equation}
The pairing function $\psi^*$ is inferred from a \emph{deterministic} ODE-based sampling procedure $\xt=\net^*_\smp(\xs)$ given a pre-trained denoising network $\net^*$. In our implementation we use the DSM described in~\citep{Tero2022EDM}, commonly known as Elucidating Diffusion Models (EDM), along with its multi-stepped deterministic sampling procedure $\net^*_\smp$ that gradually reduces the noise level $\sigma$ in $\xt+\sigma_{\max} \xs \approx \sigma_{\max} \xs $, down to a negligible level where $\xt+\sigma_{\min} \xs \approx \xt $ (details in Appendix~\ref{append:impl}). Let us discuss the desirable properties of the LMM, and further develop it.

\bf{Unambiguous Pairing. }\normalfont As elaborated in the previous section, training that ties every $\xs \sim \ps$ with every $\xt \sim \pt$ by conditioning the models on $\xt$ and marginalizing over this variable leads to unwanted detours in the flow map trajectories. The deterministic pairing we use, $\xt= \net^*_\smp(\xs)$ for every $\xs \sim \ps$, corresponds to example pairs $\xs,\xt$ that sample an implicit change-of-variable function $\xt = \psi^*(\xs)$ induced by $\net^*_\smp(\xs)$ and $\net^*$. Thus, given a state-of-the-art $\net^*$ generating samples of superior quality, the mapped distribution can be considered as a good approximation, $p_{\net_\smp^*} \approx \pt$, in this respect. Consequently, Eq.~\ref{eq:ours} regresses $\net_\theta$ under a well-defined and \emph{unambiguous pairing} between the source and target distributions \emph{regardless} of the severity of the noise level $\sigma$.

\bf{Globally Straight Trajectories. }\normalfont Assuming $\net_\theta $ is sufficiently expressive, and it satisfies Eq.~\ref{eq:ours} sufficiently well, then the lines $\xt+\sigma_t \xs$ corresponds to its iso-contours. Thus, $\net_\theta$ encodes globally straight probability flow lines connecting the source and target distributions. As noted above, while certain constructions of conditional flow maps may consist of globally straight flows, this property is lost once they are marginalized over $\xt$. In general, Eq.~\ref{eq:ours} does not pose conflicting objectives that need to be resolved.

An exception to this claim, is the availability of training pairs $ \xs,\net^*_\smp(\xs)$ and $\xs',\net^*_\smp(\xs')$ whose connecting segments do intersect and at the same time $t$, i.e., $(1-t) \xs +t\net^*_\smp(\xs) = (1-t) \xs' +t\net^*_\smp(\xs') $. In this case the regression in Eq.~\ref{eq:ours} is likely to result in an compromised intermediate solution. Such a scenario is expected to undermine both the quality of the output samples $\net_\prm$ produces, and its ability to maintain a straight iso-lines. It is important to note however, that the training examples $\xs,\net^*_\smp(\xs)$ used in~Eq.~\ref{eq:ours} correspond to solutions of a well-defined ODEs and, as discussed in~\citep{Liu2023RectFlow}, this implies that they are connected by non-intersecting smooth flow trajectories of $\net*$. 

Furthermore, the evaluation presented in Appendix~\ref{append:abl} demonstrates that the flows generated by the LMM maintain a very high degree of straightness, where only negligible improvement is made when applying more than 2 sampling steps. In addition, as shown in Section~\ref{sec:results} the samples it produces at these small NFEs receive state-of-the-art FID scores. 

Indeed this finding has motivated us to concentrate our efforts on improving the quality of the samples produced, i.e., the end-points of the lines, rather than their straightness by considering domain-specific metrics, adversarial loss, as well as fine-tuning $\net_\prm $ to the low NFE sampling schemes used in practice, as we describe below.

\bf{Constant-Speed Parameterization. }\normalfont As noted above, constructing flows with trajectories of constant direction and \emph{speed} is the key for efficient sampling, thanks to the trivialization of their integration. The noise scheduling commonly used in DDPM is known for its small progress (denoising) during its initial phase, as noted in~\citep{Liu2023RectFlow,Lipman2023flow}, and exemplifies the need for fixed speed. By training $\net_\theta$ in Eq.~\ref{eq:ours} to match endpoints, $\xt$, rather than denoising vector fields, we obtain constancy in speed by design. Specifically, at every point along the segment $\x_\sigma = \xt+\sigma \xs$, the \emph{remaining} path towards $\xt$, is given by
\begin{equation}
\label{eq:U}
    \vel(\x_\sigma,\sigma) = \net_\prm(x_\sigma, \sigma)-\x_\sigma.
\end{equation}
Hence, a constant speed parameterization with respect to $\sigma$ is given by $\vel/\sigma$ in the VE formulation that we follow. This can be used it to derive the discretization of an arbitrary number of steps in which a uniform progression along the lines is made.

\bf{Domain-Specific Loss. }\normalfont Another benefit of matching $\xt$ in Eq.~\ref{eq:ours}, rather than flow vectors, such as $\xt-\xs$ as done in~\citep{Lipman2023flow,Liu2023RectFlow}, allows us to exploit the fact that $\net_\theta$ matches native signals, and hence domain-specific metrics can be employed. In particular, this allows us to use the perceptual loss in~\citep{Johnson2016Ploss} to define $\| \cdot \|_\mathcal{P}$, when $\pt$ corresponds to a distribution of images. Indeed in Appendix~\ref{append:abl} we compare the use of this metric to $L_2$ loss, and show a substantial improvement in sampling fidelity lowering the FID over the CIFAR-10 dataset from 5.125 to 3.124 (NFE=1), and from 4.289 to 2.796 (NFE=2).

\bf{Adversarial Loss. }\normalfont Eq.~\ref{eq:ours} trains $\net_\theta$ to replicate samples $\xt$ generated by $\net^*_\smp(\xs)$, rather than being trained directly on authentic (input) samples from $\pt$. This sets a limit over the quality at which $\net_\theta$ approximates $\pt$---one which is bounded by the quality of the mediator network $\net^*$ and its sampling procedure, $\net^*_\smp$. Training $\net_\theta$ to produce signals in their original domain, e.g., clean images, in Eq.~\ref{eq:ours}, offers yet another advantage; we can follow the strategy of~\citep{Kim2024CTM} and bootstrap $\net_\theta$ to the original training data using an adversarial loss. Specifically, we train a discriminator network $\dis$ to discriminate between authentic training samples $\xt \sim \pt$ and ones produced by $\net_\theta$, by
\begin{equation}
    \mathcal{L}_\text{disc} = \Ex_{\sigma,\delta(\xt,\psi^*(\xs)),\ps(\xs)} \Big[  \log\Big( 1- \dis\big( \net_\prm\big(\xt+\sigma \xs, \sigma \big) \big) \Big) \Big]  +  \Ex_{\pt(\xt)} \Big[ \log \big( \dis (\xt) \big) \Big],
\end{equation}
where we use the architecture in~\citep{Sauer2022StyleGAN} and the adaptive weighing $\lambda_{\text{adapt}}$ in~\citep{Esser2O21VQGAN} that Kim et al. use. We finally train $\net_\theta$ to minimizing $\lambda_{\text{lines}} \mathcal{L}_\text{lines} + \lambda_{\text{adapt}} \mathcal{L}_\text{disc} $. We provide all the implementation details in Appendix~\ref{append:impl}.

    

As we show in Appendix~\ref{append:abl}, training the LMM without $ \mathcal{L}_\text{disc}$ achieves fairly satisfying FID scores, namely 3.12 (NFE=1) and 2.79 (NEF=2) on CIFAR-10. By incorporating the latter these scores further improve to 1.67 (NFE=1) and 1.39 (NFE=2). This \emph{surpasses} the quality of samples generated by $\net^*_\smp(\xs)$ which uses more sampling steps (NFE=35) and achieves FID of 1.79. 

\bf{Sampling-Optimized Training (SOT). }\normalfont Motivated by the evaluation reported in Appendix~\ref{append:abl}, showing that the LMM achieves its high-quality samples already at NFE $\leq 3$, we explored the option of further improving its performance by restricting the training of $\net_\prm$ to the specific steps (noise levels $\sigma$) used at sampling time. Appendix~\ref{append:abl} shows the further quality increase, from FID 1.67 to 1.57 (NFE=1), thanks to this training strategy.

\section{Evaluation and Comparison}
\label{sec:results}

\begin{table}
\vspace{-0.07in}
\centering
\renewcommand{\arraystretch}{1.0} 
\tiny
\setlength{\tabcolsep}{4pt} 
\begin{minipage}[t]{0.5\textwidth}
\centering
\caption{CIFAR-10}
\label{tab:cifar}
\begin{tabular}{@{}l c c c c@{}}
\toprule
\multirow{2}{*}{Model} & \multirow{2}{*}{\textbf{NFE}} & \multicolumn{2}{c}{ unconditional} & \multicolumn{1}{c}{conditional}\\ 
\cmidrule(lr){3-4} \cmidrule(lr){5-5}
& & \textbf{FID} & \textbf{IS} & \textbf{FID} \\ \midrule
\textbf{GAN} \\
BigGAN~\cite{biggan} & 1 & 14.70 & 9.22 & - \\
StyleGAN2-ADA~\cite{StyleGANADA} & 1 & 2.92 & 9.83 & 2.42 \\
StyleGAN-D2D~\cite{StyleGAND2D} & 1 & - & - & 2.26 \\
StyleGAN-XL~\cite{Sauer2022StyleGAN} & 1 & - & - & 1.85 \\ \midrule
\textbf{Diffusion / Score Matching} \\
DDPM~\cite{Ho2020DDPM} & 1000 & 3.17 & 9.46 & - \\
DDIM~\cite{Song2021implicit} & 100 & 4.16 & - & - \\ 
Score SDE~\cite{Song2021SMLD} & 2000 & 2.20 & 9.89 & - \\ 
EDM~\cite{Tero2022EDM} & 35 & 1.97 & 9.84 & 1.79  \\ \midrule
\textbf{Distillation / Direct Gen.} \\
KD~\cite{Luhman2021knowdistl} & 1 & 9.36 & 8.36 & - \\
PD~\cite{Salimans2022PD} & 1 & 9.12 & - & - \\
CT~\cite{Song23CM} & 1 & 8.70 & 8.49 & - \\
CD~\cite{Song23CM} & 1 & 3.55 & 9.48 & - \\
CD+GAN~\cite{Lu2023CMGAN} & 1 & 2.65 & - & - \\
iCT~\cite{song2024improved} & 1 & 2.83 & 9.54 & - \\
iCT-deep~\cite{song2024improved} & 1 & 2.51 & 9.76 & - \\
CTM~\cite{Kim2024CTM} & 1 & 1.98 & - & 1.73 \\
DMD~\cite{yin2024onestepdmd} & 1 & 3.77 & - & 2.66 \\
SiD ($\alpha=1$) ~\cite{Zhou2024SiD} & 1 & 2.02 & 10.02 & 1.93 \\
SiD ($\alpha=1.2$) ~\cite{Zhou2024SiD} & 1 & 1.92 & 9.98 & 1.71 \\
PD~\cite{Salimans2022PD} & 2 & 4.51 & - & - \\
CT~\cite{Song23CM} & 2 & 5.83 & 8.85 & - \\
CD~\cite{Song23CM} & 2 & 2.93 & 9.75 & - \\
iCT~\cite{song2024improved} & 2 & 2.46 & 9.80 & - \\
iCT-deep~\cite{song2024improved} & 2 & 2.24 & 9.89 & - \\
CTM~\cite{Kim2024CTM} & 2 & 1.87 & - & 1.63 \\ \midrule
\textbf{Flow Matching} \\
OT-CFM~\cite{Lipman2023flow} & 142 & 6.35 & - & - \\
1-Rect-Flow (distill)~\cite{Liu2023RectFlow} & 1 & 6.18 & 9.08 & - \\
2-Rect-Flow (distill)~\cite{Liu2023RectFlow} & 1 & 4.85 & 9.01 & - \\
3-Rect-Flow (distill)~\cite{Liu2023RectFlow} & 1 & 5.21 & 8.79 & - \\
1-Rect-Flow~\cite{Liu2023RectFlow} & 127 & 2.58 & 9.60 & - \\
2-Rect-Flow~\cite{Liu2023RectFlow} & 110 & 3.36 & 9.24 & - \\
2-Rect-Flow~\cite{Liu2023RectFlow} & 104 & 3.96 & 9.01 & - \\ \midrule
LMM  & 1 & \textbf{1.90} & \textbf{10.16} & \textbf{1.57}\normalfont \\
LMM  & 2 & \textbf{1.55} & \textbf{10.20} & \textbf{1.39}\normalfont \\ \bottomrule
\vspace{0.1in}
\end{tabular}

\begin{tabular}{c}
    \setlength{\tabcolsep}{9pt}
    \begin{tabular}{ccccc}
         \multirow{2}{*}{\textbf{EDM}} & \multicolumn{4}{c}{\textbf{LMM}} \\\cmidrule(lr){2-5} 
            & \multicolumn{2}{c}{ wo/ADL} & \multicolumn{2}{c}{ w/ADL} \\ \cmidrule(lr){2-3}\cmidrule(lr){4-5}
         NFE 79 & NFE 1 & NFE 2 & NFE 1 & NFE 2 \\
    \end{tabular} \vspace{0.02in} \\
    \includegraphics[width=2.3in]{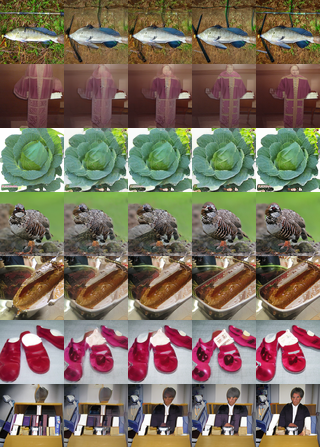}
    \end{tabular}
    
    \caption{ImageNet \rsm~Samples Comparison.}
    \label{tab:visual}
\end{minipage}
\hfill
\begin{minipage}[t]{0.45\textwidth}
\centering
\caption{ImageNet \rsm}
\label{tab:imagenet}
\begin{tabular}{@{}l c c c@{}}
\toprule
 \multirow{2}{*}{Model} & \multirow{2}{*}{\textbf{NFE}}  & \multicolumn{2}{c}{conditional}  \\  \cmidrule(l){3-4}
 & & \textbf{FID} & \textbf{IS} \\\midrule
\textbf{GANs} \\
BigGAN-deep~\cite{biggan} & 1 & 4.06 & - \\
StyleGAN-XL~\cite{Sauer2022StyleGAN} & 1 & 1.51 & \textbf{82.35} \\ \midrule
\textbf{Diffusion / Score Matching} \\
RIN~\cite{Jabri23RIN} & 1000 & 1.23 & - \\
EDM~\cite{Tero2022EDM} & 511 & 1.36 & - \\
DDPM~\cite{Ho2020DDPM} & 250 & 11 & - \\
EDM~\cite{Tero2022EDM} & 79 & 2.23 & 48.88 \\ \midrule
\textbf{Distillation / Direct Gen.} \\
PD~\cite{Salimans2022PD} & 1 & 15.39 & - \\
BOOT~\cite{gu2023boot} & 1 & 16.30 & - \\
CT~\cite{Song23CM} & 1 & 13.0 & - \\
CD~\cite{Song23CM} & 1 & 6.20 & 40.08 \\
iCT~\cite{song2024improved} & 1 & 4.02 & - \\
iCT-deep~\cite{song2024improved} & 1 & 3.25 & - \\
CTM~\cite{Kim2024CTM} & 1 & 1.92 & 70.38 \\
DMD~\cite{yin2024onestepdmd} & 1 & 2.62 & - \\
SiD ($\alpha=1$) ~\cite{Zhou2024SiD} & 1 & 2.02 & - \\
SiD ($\alpha=1.2$) ~\cite{Zhou2024SiD} & 1 & 1.52 & - \\
PD~\cite{Salimans2022PD} & 2 & 8.95 & - \\
CT~\cite{Song23CM} & 2 & 11.1 & - \\
CD~\cite{Song23CM} & 2 & 4.70 & - \\
iCT~\cite{song2024improved} & 2 & 3.20 & - \\
iCT-deep~\cite{song2024improved} & 2 & 2.77 & - \\
CTM~\cite{Kim2024CTM} & 2 & 1.73 & 64.29 \\ \midrule
\textbf{Flow Matching} \\
OT-CFM~\cite{Lipman2023flow} & 138 & 14.45 & - \\
BOT-CFM~\cite{Pooladian2023mini} & 132 & 11.82 & - \\ \midrule
LMM & 1 & \textbf{1.47} & 59.86 \\ 
LMM & 2 & \textbf{1.17} & 61.18 \\ 
\bottomrule
\end{tabular}

\vspace{0.5cm}

\caption{AFHQ \rsm}
\label{tab:afhq}
\begin{tabular}{@{}l c c @{}}
\toprule
Model & \textbf{NFE} & \textbf{FID}\\ \midrule
\textbf{Score Matching} \\
EDM~\cite{Tero2022EDM} & 79  & 1.96 \\ \midrule
\textbf{Distillation} \\
SiD ($\alpha=1.2$) ~\cite{Zhou2024SiD} & 1 & 1.71  \\ 
SiD ($\alpha=1$) ~\cite{Zhou2024SiD} & 1 & \textbf{1.63}  \\ \midrule
LMM & 1  &  2.68 \\
LMM & 2  &  \textbf{1.54}  \\ \bottomrule

\end{tabular}
\end{minipage}

\normalsize
\end{table}

We trained the LMM on three benchmark datasets, CIFAR-10, ImageNet \rsm, and AFHQ \rsm, which are commonly used for evaluating generative models. We used the same network architecture and hyper-parameters as existing models, with all the implementation details provided in Appendix~\ref{append:impl}.

\bf{Quantitative Comparison. } \normalfont Table~\ref{tab:cifar} provides a comprehensive comparison of the CIFAR-10 reproduction quality achieved by different models. The comparison clearly shows that diffusion-based models achieve lower FID scores, albeit at an increased sampling cost compared to GANs. Flow-matching models demonstrate their ability to reduce the NFEs, alongside a range of distillation techniques that operate effectively with very low NFEs---one or two sampling steps.

Among these, the Consistency Trajectory Model (CTM)~\citep{Kim2024CTM}, achieves excellent FID scores of 1.73 (NFE=1) and 1.63 (NFE=2) on conditional CIFAR-10. Our LMM surpasses these scores and sets new state-of-the-art scores of 1.57 and 1.39 respectively. We note that both methods benefit from the use of an adversarial loss, but as reported in Section~\ref{append:abl}, the LMM's performance remains better also without this loss. We attribute this to the fact that the LMM produces favorable line flow trajectories, rather than relying on the curved EDM trajectories that the CTM distills.

The Rect-Flow in~\citep{Liu2023RectFlow} achieves an impressive FID score of 4.85 in its second iteration, where it produced significantly straighter trajectories (as shown in Table~\ref{fig:traj}). We note that this second iteration achieves the best trade-off between straightness and errors that this scheme accumulates.

Table~\ref{tab:imagenet} shows the results obtained on a larger dataset, ImageNet \rsm. Here too, the LMM demonstrates state-of-the-art performance, with a notable improvement at NFE=2, where it reaches an FID of 1.17. The SiD~\citep{Zhou2024SiD} trains a single-step generator to agree with a pre-trained EDM, achieving an impressive score of 1.52. Unlike the LMM, SiD does not rely on the EDM to generate training examples; instead, it uses it to define the generator's loss while simultaneously training an additional score-matching network. This approach poses significantly higher GPU memory requirements and operations during training.

Finally, Table~\ref{tab:afhq} reports the results on the AFHQ \rsm~dataset, where the LMM shows lower FID scores using significantly fewer NFEs compared to the EDM despite the fact that the latter is used to produce the initial correspondence between $\ps$ and $\pt$. This is also the case in Tables~\ref{tab:cifar} and \ref{tab:imagenet}. While achieving a state-of-the-art FID of 1.54 at NFE=2, the SiD achieves a better score using a single step. We note that unlike the CIFAR-10 and ImageNet \rsm cases, the discriminator architecture and hyper-parameters we used were not we used were not tailored to this dataset in previous work (e.g., StyleGAN-XL~\citep{Sauer2022StyleGAN}). This affected the expected improvements from the SOT strategy, as discussed in Appendix~\ref{append:abl}, and we therefore believe the LMM has greater potential on this dataset.

In terms of Inception Score (IS), the LMM achieves state-of-the-art results, scoring above 10 for both NFEs on CIFAR-10, as shown in Table~\ref{tab:cifar}. On ImageNet \rsm, the LMM improves upon its teacher model (EDM), although StyleGAN-XL attains the highest score. Among diffusion-based models, the LMM receives an IS of 61.18 using 2 NFEs, which is closely competitive with CTM which scores 64.29.

\bf{Visual Evaluation. } \normalfont Table~\ref{tab:visual} compares samples produced by the EDM and LMM with and without an Adversarial Loss (ADL), and using 1 or 2 sampling steps. Incorporating the ADL appears to be related to fine image details, contributing to their richness and resolvedness. This is observed in the fish background, lettuce leaves, the bird feathers, and the man's face. The second sampling iteration (NFE=2 in the table) has a larger scale impact, improving the correctness of the objects' shape, as well as the consistency between different objects. This effect can be seen in the clerk's body and face, the bird's body, the shape of the bread/cake, and the matching red shoes. The images generated by the EDM generally appear to be less detailed, although there are clear exceptions to that. A similar comparison of CIFAR-10 and AFHQ \rsm~is shown in Table~\ref{tab:comp2}.

\begin{figure}
    \centering
     \includegraphics[width=4.5in]{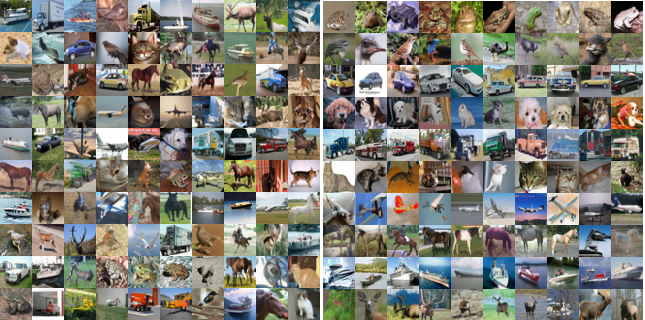}
    \caption{LMM Generated CIFAR-10 Samples. Class unconditional on the left, and conditional on the right. Rows correspond to different classes.}
    \label{fig:samps_cifar}
\end{figure}

\begin{figure}
    \centering
     \includegraphics[width=2.5in]{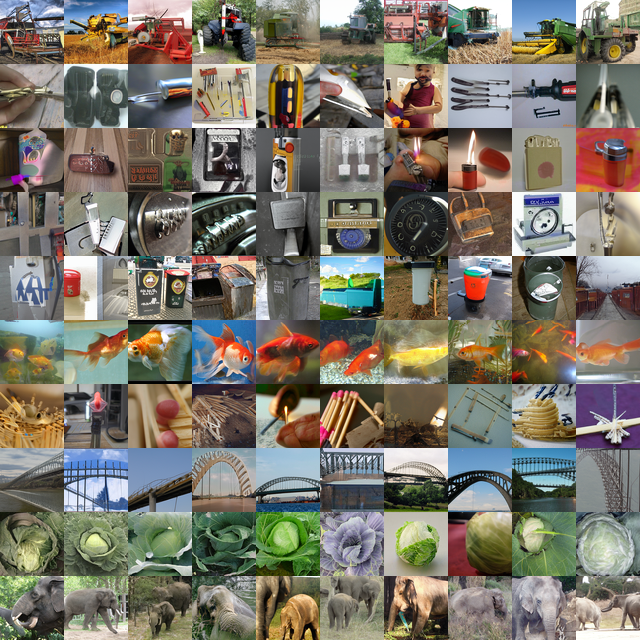}
    \caption{LMM Generated Conditional ImageNet \rsm~Samples. Rows correspond to different classes.}
    \label{fig:samps_imagenet}
\end{figure}

\begin{figure}
    \centering
     \includegraphics[width=2.5in]{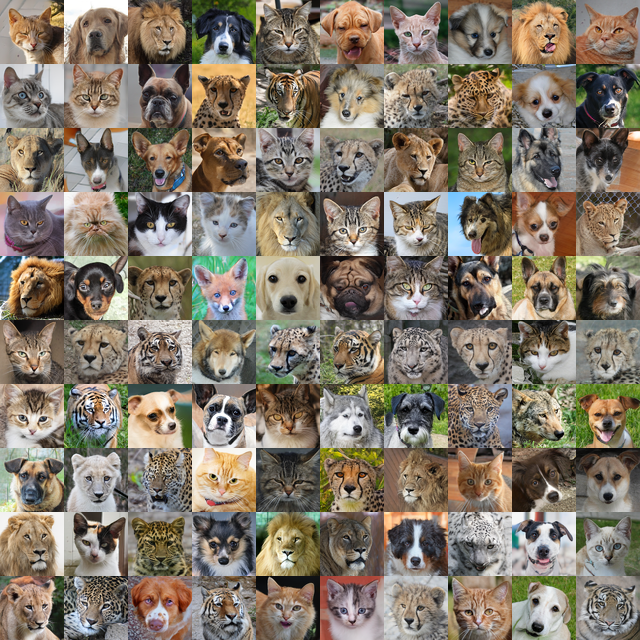}
    \caption{LMM Generated AFHQ \rsm~Samples.}
    \label{fig:samps_afhq}
\end{figure}

Additional example samples produced by the LMM on each of the datasets are shown in Figures~\ref{fig:samps_cifar}, ~\ref{fig:samps_imagenet}, and~\ref{fig:samps_afhq}.
\section{Conclusions}
\label{sec:conc}

In this work, we showed that using broad random correspondences between the source and target distributions results in collapsed predictions at low SNRs. By bringing them to a common analytical framework, we showed that this degeneracy is inherent to key models, including denoising diffusion, score-matching, and flow-matching techniques. We used this insight to propose a solution by deriving a deterministic correspondence from ODE-based sampling. To avoid the inefficiencies in the resulting trajectories, we only used their endpoints to train our LMM which parameterizes the transition between distributions using globally straight lines.

We leveraged the fact that our formulation works directly on signal reconstruction, and proposed several training losses and strategies to improve the quality of the generated samples. In doing so, our work bridges the domains of flow matching and denoising distillation. The combined effect of enhanced sampling quality and sampling efficiency has enabled the LMM to achieve state-of-the-art image generation quality in just one or two sampling steps.

Finally, as part of our effort to understand and improve the pairing required for training flow models, we made a theoretical contribution showing that OT-based pairing suffers from an exponential relationship between the size of the paired sets (mini-batches) and the signal dimension.

Our work leaves one important goal unaddressed: avoiding the reliance on a pre-trained model, and establishing its pairing in an ab initio manner. As a future research direction we intend to investigate the adaptation of an iterative scheme, like the one in~\cite{Liu2023RectFlow}, while avoiding drifts in the training data during this process.

\bf{Code Reproducibility Statement. } \normalfont In Appendix~\ref{append:impl} we provide detailed information on the network architecture and hyper-parameters used to produce the reported results. Additionally, we plan to publicly release our code and the trained LMM network weights.

\bf{Social Impact Statement. } \normalfont Given their increasing prevalence, improving the efficiency of generative AI models is likely to result in a significant reduction in computational costs and energy usage. However, we are fully aware of the risks associated with these models and wish to express our strong opposition to any unethical use.


\bibliography{iclr2025_conference}
\bibliographystyle{iclr2025_conference}

\appendix

\section{Appendix}

\subsection{Regression at low signal-to-noise Ratios}
\label{append:reg}

Both DDPM~\citep{sohl2015thermo,Ho2020DDPM} and DSM, e.g., NCSN~\citep{Song2021NCSN} and EDM~\citep{Tero2022EDM}, start their sampling process from an easy-to-sample source distribution $\x^N \sim \ps$, typically a Gaussian. Hence their denoising networks $\net_\prm$ are trained to operate on these distributions. At $t=N$, Eq.~\ref{eq:denoising} becomes
\begin{equation}
\label{eq:denoising_zero}
\text{argmin}_\prm \Ex_{q(\xt),p(\x|\xt,N)} \Big[  \Vert \net_\prm(\x,s_N) - \nabla_{\x}\log p(\x|\xt,N) \Vert^2 \Big].
\end{equation}
In the case of NCSN, $ s_N = \sigma_N$ and $p(\x|\xt,\sigma_N) = \nrml (\xt,\sigma_N^2 I)$ where $\sigma^2_N >> \Vr [\xt] $, i.e., a very low SNR, allowing to be approximate this distribution by a pure noise at sampling time, specifically $\ps = \nrml (0,\sigma^2_N I) $. 

Noting that $\nabla_{\x}\log p(\x|\xt,N) = (\xt-\x) / \sigma_N^2$, Eq.~\ref{eq:denoising_zero} becomes (at $t=N$),
\begin{equation}
\label{eq:SMN}
 \text{argmin}_\prm \Ex_{q(\xt), \xs \sim \ps } \Big[  \Vert \net_\prm(\xs,\sigma_N) - (\xt - \xs)/\sigma_N^2 \Vert^2 \Big],
\end{equation}
where every $\xs \sim \ps$ is equally regressed to match every $\xt \sim \pt$. Regressing under such indeterminacy is poised to result in the degenerate averaged prediction $\net_\prm(\x,\sigma_N) = (\mu_\pt -\x)\sigma_N^{-2} $. This topic is thoroughly discussed in~\citep{kendall1973book,Allan2013noisefit}. Finally, we note that at sampling stage the factor $\sigma_N^{-2}$ is typically canceled by using time steps proportional to $\sigma_N^2$, see for example~\citep{Song2021NCSN} and~\citep{Tero2022EDM}. Thus, the sampling trajectories are drawn towards $\mu_\pt$, up to some implementation-dependent speed factors, during their first steps. This effect is highlighted by the green arrows in Table~\ref{fig:traj}.

Analogously, the DDPM noise scheduling is set such that $\alpha_N $ is small, e.g., $\alpha_N=6 \times 10^{-3}$ in~\citep{Ho2020DDPM} and $\alpha_N=5 \times 10^{-5}$ in~\citep{Nichol21improved}. Therefore $p(\x|\xt,N) = \nrml (\sqrt{\alpha_N}\xt,(1-\alpha_N) I) \approx \nrml (0,I) $ which, here as well, can be replaced with the source distribution $\ps$ during sampling. In this case, Eq.~\ref{eq:denoising_zero} becomes (again, at $t=N$),
\begin{equation}
\label{eq:DDPMN}
\text{argmin}_\prm \Ex_{q(\xt), \xs \sim \ps} \Big[  \Vert \net_\prm(\xs,N) - (\sqrt{\alpha_N} \xt - \xs) / (1-\alpha_N) \Vert^2  \Big],
\end{equation}
resulting in $ \net_\prm(\x,N) = (\sqrt{\alpha_N} \mt -\x)/(1-\alpha_N) $. In fact, this should be interpreted as $ \net_\prm(\x,N) = (\sqrt{\alpha_N} \mt  + \sqrt{(1-\alpha_N)} \ms -\x)/(1-\alpha_N) \approx \ms -\x $, since $\alpha_N<<1$ and we added the term $\sqrt{(1-\alpha_N)} \ms$ since $\ms=0$. More fundamentally, the DDPM noising process $p(\x|\xt,N) = \nrml (\sqrt{\alpha_t}\xt,(1-\alpha_t) I) $ gradually replaces every data sample $\xt$ with a normal Gaussian by shifting the mean from $\xt$ towards $\ms $  (chosen to be 0 for convenience) and by increasing the variance from $(1-\alpha_1) \approx 0 $ to $\Vr[\xs] = 1$.

Thus, at the $N$-th step of the DDPM sampling step, the denoiser collapses to the mean of the source distribution. Consequently, its flow trajectories gravitate toward $\ms=0$ during their earlier steps. This affects only the magnitude of initial (full noise) states, and the sample's shape evolves only in later steps, thus the DDPM sampling process is often described as stagnant during its early stages (e.g., in Figure 6 in~\citep{Lipman2023flow}).



Finally, the flow models in~\citep{Lipman2023flow}, and~\citep{Liu2023RectFlow} at 1-Rect-Flow, regress arbitrary samples from $\ps$ to the data points $\xt$ at time $t=0$, where its training loss, Eq.~\ref{eq:fsf}, becomes
\begin{equation}
\label{eq:fsf}
\text{argmin}_\prm \Ex_{q(\xt),p(\xs)} \Big[ \Vert \net_\prm\big(\xs,0 \big) - (\xt-\xs) \Vert^2 \Big],
\end{equation}
Similarly to Eq.~\ref{eq:SMN}, also Eq.~\ref{eq:fsf} regresses points $\xs$ with direction towards arbitrary data points, $\xt-\xs$. This leads again to a degenerate solution where $\net_\prm(\xs,0) = \mt-\xs$, which similarly to the score-matching approach, biases the sampling trajectories towards $\mt$ at their earlier stages. This effect is also observed in Table~\ref{fig:traj}.

\subsection{Batch Optimal Transport - Batch Size Analysis}
\label{append:OTB}

We assess here the asymptotic dependence in the Batch OT CFM (BOT-CFM) methods described in~\citep{Pooladian2023mini} and~\citep{Tong2024mini} over the batch size $\bs$ as a function of space dimension $\sdim$. In these works, following the notations of the former, the independent distribution $ \ps(\xs)\pt(\xt) $ in Eq.~\ref{eq:cfm} is replaced by a joint distribution $q(\xs,\xt)$ induced by a batch-optimized coupling, $\{\xs^{i}\}_{i=1}^\bs \sim \ps $ and $\{\xt^{j_i}\}_{i=1}^\bs \sim \pt $, where the permutation $j_i$ optimizes the transport cost $\| \xs^{i} - \xt^{j_i}\|^2$ within each batch. Combining this with the OT conditional flow map $\psi_{\xt}(\x,t)$ in~\citep{Lipman2023flow}, the BOT-CFM training loss is given by
 \begin{equation}
 \label{eq:CFMBOT}
   \text{argmin}_\prm \Ex_{t,\{\xs^{i}\}_{i=1}^\bs \sim \ps,\{\xt^{i}\}_{i=1}^\bs \sim \pt} \Big[ \sum_{i=1}^\bs \Vert \net_\prm \big( (1- t) \xs^{i} + t \xt^{j_i},t\big)  - \big(\xt^{j_i} - \xs^{i} \big)    \Vert ^2 \Big],
 \end{equation}
 
To simplify the analysis we consider a fairly naive problem of finding a mapping from a normal Gaussian in $ \Rn^\sdim$ to itself, where the optimal solution is given by the identity mapping. In the context of matching the velocity field, as done in~\citep{Pooladian2023mini,Tong2024mini}, the optimal field is given by $ \net_\prm(\x,t)=0$. As shown in Appendix~\ref{append:reg}, in case of independent distribution $ \ps(\xs)\pt(\xt) $ (the solution of Eq.~\ref{eq:fsf})) the resulting vector field at $t=0$ is $\net_\prm(\xs,0) = \mt-\xs=-\xs \neq 0$ which is clearly far from the optimum. 

In the BOT-CFM (at $t=0$) closer and closer $\xt^{j_i}$ will be found to each $\xs^i$ as the batch size increases, and hence by training $\net_\prm(\xs^i,0)$ to match $ \xt^{j_i} - \xs^{i} $, in Eq.~\ref{eq:CFMBOT}, a reduced velocity vector is expected. The question of how fast this decrease takes place as a function of $\sdim$ is critical, as only moderately sized batches can be used in practice.

We address this question at $t=0$, where Eq.~\ref{eq:CFMBOT} simplifies to a simple regression problem over $\xs$,
 \begin{equation}
 \label{eq:CFMBOT0}
   \text{argmin}_\prm \Ex_{\{\xs^{i}\}_{j=1}^\bs \sim \ps,\{\xt^{i}\}_{i=1}^\bs \sim \pt} \Big[ \sum_{i=1}^\bs \Vert \net_\prm \big(  \xs^{i} ,0\big)  - \big(\xt^{j_i} - \xs^{i} \big) \Vert ^2 \Big],
\end{equation}
which is solved by,
\begin{equation}   
\label{eq:sol0}
\net_\prm \big(  \xs ,0\big) = \Ex_{p^{\batch^\bs}(\xt^*|\xs)} \big[ \xt^*  -\xs \big],
\end{equation}
where $p^{\batch^\bs} (\xs,\xt^*)$ is the joint distribution induced by finding the optimal pairing between source $\xs^{i}$ and target $\xt^{j_i}$ within each batch $\batch^\bs$ of size $\bs$. 

The case $\bs\!=\!1$ (equivalent to random pairing), we get $p^{\batch^1}(\xt^*|\xs) = \ps(\xs)\pt(\xt)$ which was discussed above and results in a velocity $\net_\prm (\xs,0) =  -\xs$ attracting sampling trajectories towards $\mt \!=\! 0$ at $t\!=\!0$, instead of remaining stationary, thus producing the unnecessarily curved trajectories. As $\bs$ increases, however, the chances to regress $\xs$ to closer $\xt^*$ increases and thus a shift in $ \Ex_{ p^{\batch^\bs}(\xt^*|\xs)} \big[ \xt^* \big] $ toward $\xs$ is expected. In order to analyze the magnitude of this shift as a function of both $\bs$ and $\sdim$, let us review basic properties of random vectors in $\Rn^\sdim$.

Let $x$ and $y$ be two independent normal scalars drawn from $\mathcal{N}(0,1) $. Their product $xy$ is a random variable with the following moments
\begin{equation}
\label{eq:mean}
    \Ex [ xy ] =  \Ex [ x ]  \Ex [ y ]  = 0
\end{equation}
and,
\begin{equation}
\begin{aligned}
\label{eq:var}
\Vr & [ xy ]  =  \Ex\big[ (xy)^2\big] =  \Ex\big[ x^2 \big] \Ex\big[ y^2 \big] = \Vr [ x ]\Vr [ y ]= 1 < \infty,
\end{aligned}
\end{equation}
both follow from the normality and independence of $x,y$. Let us consider now two independent normal vectors $\x,\y \in \Rn^\sdim$, drawn from $\mathcal{N}(0,I) $, and their dot-product, defined by
\begin{equation}
    \dotp{\x}{\y} = \frac{1}{\sdim} \sum_{i=1}^\sdim \x^i \y^i.
\end{equation}
Being an average of independent random variables, at large space dimension $\sdim$ the central limit theorem becomes applicable and provides us its limit distribution by,
\begin{equation}
\label{eq:clt}
\dotp{\x}{\y} \stackrel{\sdim}{\rightarrow} \mathcal{N}(0,\sdim^{-1}),
\end{equation}
which is calculated from the scalar moments in Eq.~\ref{eq:mean} and Eq.~\ref{eq:var}. This implies that as the space dimension $\sdim$ increases, this distribution gets more concentrated around 0, meaning that the vectors $\x$ and $\y$ are becoming less likely to be related to one another by becoming increasingly orthogonal. As we shall now show, this makes the task of finding $\xt \in \batch^\bs$ close to $\xs$ within finite batches increasingly difficult as $\sdim$ grows. This relates to a well-known phenomenon where the ratio between the farthest and closest points converges to a constant, as the space dimension increases~\citep{Beyer1999NN}.

Indeed, by considering the magnitude of the regressed flow velocity in Eq.~\ref{eq:norm},
\begin{equation}
\label{eq:norm}
\begin{aligned}
     & \bigl\|  \Ex_{ p^{\batch^\bs}(\xt^*|\xs)} \big[ \xt^* \big] -\xs \bigr\| ^2=\bigl\| \Ex_{ p^{\batch^\bs}(\xt^*|\xs)} \big[ \xt^* \big]  \bigr\| ^2 + \| \xs \|^2 - 2 \Dotp{\Ex_{ p^{\batch^\bs}(\xt^*|\xs)} \big[ \xt^* \big]}{\xs}\\
     & \geq   \| \xs \|^2 - 2 \Dotp{\Ex_{ p^{\batch^\bs}(\xt^*|\xs)} \big[ \xt^* \big]}{\xs} =  \| \xs \|^2 - 2 \Ex_{ p^{\batch^\bs}(\xt^*|\xs)} \big[ \dotp{ \xt^* }{\xs}\big] ,
\end{aligned}
\end{equation}
we clearly see the need for increased dot-product similarity within the batches $\batch^\bs$ in order to reduce the magnitude of the learned target flow velocity---ideally zero in this problem. In this derivation $\| \Ex_{ p^{\batch^\bs}(\xt^*|\xs)} \big[ \xt^* \big]  \| ^2$ is neglected as we are in a process of deriving a lower bound for the flow velocity field, $\| \Ex_{ p^{\batch^\bs}(\xt^*|\xs)} \big[ \xt^* \big] -\xs\|^2$. We also note that the last equality follows from the linearity of the dot-product operator. 

As an upper bound for $\dotp{\Ex_{ p^{\batch^\bs}(\xt^*|\xs)} \big[ \xt^* \big]}{\xs}$ we assume that this similarity is computed by pairing $\xs$ with its \emph{closest} $\xt^* \in \batch^\bs$ without considering trade-offs that arise when pairing a complete batch of source points $\{\xs^{i}\}_{j=1}^\bs \sim \ps$ with the batch of target points, in $\batch^\bs$, as done in practice in BOT-CFM, in Eq.~\ref{eq:CFMBOT}. 

In this scenario, $\dotp{\xt^*}{\xs} = \max_i \dotp{\xt^i}{\xs}$, where $\dotp{\xt^i}{\xs}$ are independent variables and, as shown above, $\dotp{\xt^i}{\xs} \sim \mathcal{N}(0,\sdim^{-1})$. Using Jensen's inequality, we get that
\begin{equation}
\label{eq:mag}
\begin{aligned}
\exp \big(t \Ex_{ p^{\batch^\bs}(\xt^*|\xs)} [\dotp{\xt^*}{\xs}]\big) & \leq \Ex_{ p^{\batch^\bs}(\xt^*|\xs)} \big[ \exp(t \dotp{\xt^*}{\xs}) \big] = \Ex_{ \mathcal{N}(0,\sdim^{-1}) }\big[ \max_i  \exp(t \dotp{\xt^i}{\xs}) \big]\\ &  \leq \sum_{i=1}^\bs \Ex_{ \mathcal{N}(0,\sdim^{-1})} \big[\exp(t \dotp{\xt^i}{\xs}) \big]  =\bs \exp\left(\frac{t^2}{2\sdim}\right), 
\end{aligned}
\end{equation}
where the last equality follows from the calculation of the moment generating function of the Gaussian distribution, $\mathcal{N}(0,\sdim^{-1})$. Thus, by taking the logarithm of Eq.~\ref{eq:mag} and dividing by $t$ we get
\begin{equation}
    \Ex_{ p^{\batch^\bs}(\xt^*|\xs)} [\dotp{\xt^*}{\xs}] \leq \log(n)/t + \frac{t}{2\sdim}.
\end{equation}
Finally, by setting $t=\sqrt{2  \sdim \log \bs}$, we get
\begin{equation}
    \Ex_{ p^{\batch^\bs}(\xt^*|\xs)} [\dotp{\xt^*}{\xs}] \leq \sqrt{\frac{2\log n}{\sdim}}.
\end{equation}
This relation implies that in order to obtain a proper (zero) target velocity field in~Eq.~\ref{eq:norm}, the batch size $\bs$ must grow exponentially as a function of the space dimension $\sdim$, which tends to be fairly large in practical settings. Indeed, as demonstrated in Table~\ref{fig:traj} already at $\sdim=128$ the BOT-CFM shows a moderate reduction in the average trajectory straightness compared to the CFM using batch sizes of $\bs=128$. The use of $\bs=256$ offered a negligible improvement. We conclude that this dependence undermines the prospect of achieving additional substantial improvement over the one reported in~\citep{Pooladian2023mini} by increasing the batch size and relying solely on the BOT strategy.

Several notes on the scope of our analysis which considered a simple problem of mapping two Gaussians and considered the affairs at $t=0$. First, it shows that even over an arguably simple problem the effectiveness of the BOT-CFM is limited by its asymptotic. Second, as discussed at great length in Section~\ref{sec:background} a major source of sampling inefficiency, shared by multiple key approaches, takes place at the vicinity of $t=0$, and hence the focus of our analysis to this time should not necessarily be considered as a limitation. Finally, most of the arguments made above remain valid when real-world target data distribution $\pt$ is used. Namely, the limiting orthogonal distribution in Eq.~\ref{eq:clt} and hence the exponential batch size requirement for finding real-world data point $\xt^*$ sufficiently close to a random latent vector $\xs~\sim \mathcal{N}(0,\sdim^{-1})$. Our restriction to a target Gaussian distribution is made specifically for the purpose of being able to consider the analytical results with respect to a \emph{known} optimal flow field.

\subsection{Ablation Studies}
\label{append:abl}

We report here the results of several empirical experiments that assess the impact of different components related to LMM's training, described in Section~\ref{sec:method}, on its sampling performance and quality.

\bf{Domain-Specific versus L2 Loss. } \normalfont Training the LMM to reproduce the end-points of the probability flow lines, i.e., noise-free images, allows us employ perceptual metrics, specifically~\citep{Johnson2016Ploss}, for training. This loss is known to provide visually-preferable optimization trade-offs in various applications, see~\citep{zhang2018perceptual}. Table~\ref{tab:abl_cifar} shows that training the LMM using a VGG-based perceptual loss (VGG) achieves lower FID scores compared to that of L2 loss at all NFEs tested. The ability to use this reconstruction loss is inherent to the design of the LMM, and is not shared by all flow-based approaches, e.g.,~\citep{Lipman2023flow,Liu2023RectFlow}.

\bf{Number of Sampling Steps. } \normalfont Tables~\ref{tab:abl_cifar}, \ref{tab:abl_imagenet}, and~\ref{tab:abl_afhq} report the FID scores on different datasets using different NFEs and sampling steps. Specifically, we used subsets of the sampling steps from the sampling scheme in~\citep{Tero2022EDM}. While the number of steps provides some amount of ability to trade-off between quality and efficiency, it is clear from these tables that increasing the NFEs suffers from a diminishing return. This finding aligns with the explanation that the probability flow lines generated by the LMM are fairly straight, and that the sampling errors are primarily due to the accuracy of their endpoints, i.e., the quality at which the target samples $\xt \sim \pt$ can be reproduced by exact integration. This further motivated us in Section~\ref{sec:method} to focus on improving the sample reproduction, as we evaluate next.

\bf{Adversarial Loss. } \normalfont Indeed, Tables~\ref{tab:abl_cifar}, \ref{tab:abl_imagenet}, and~\ref{tab:abl_afhq}, show that the incorporation of an adversarial loss (ADL) provides an additional significant improvement to the image quality produced by the LMM. Indeed, this addition also helped the CTM in~\citep{Kim2024CTM} to improve their baseline, specifically, FID of 2.28 using a discriminator and 5.19 without it, using NFE=1 on CIFAR-10. We attribute the lower FID scores achieved by the LMM, in both scenarios, to the fact that it models favorable line flow trajectories, rather than the original curved EDM's trajectories, which are distilled in~\citep{Kim2024CTM}.

\bf{Sampling-Optimized Training. } \normalfont Motivated by limited improvement higher NFEs produce, in Section~\ref{sec:method} we proposed another strategy to improve sample quality by restricting the training to the specific time steps used at the sampling stage. Tables~\ref{tab:abl_cifar} and \ref{tab:abl_imagenet} show that this training strategy also has the ability to contribute significantly despite the fact that it adds no cost. Table~\ref{tab:abl_afhq} an opposite trend which appears to be related to a saturation (over-fitting) due to two factors: (i) to limited data available in this dataset, and (ii) the SOT focuses on high noise levels, which makes it easier to discriminate between generated and real samples. We conclude that a more fine-tuned discriminator setting is needed to achieve optimal results.

\subsection{Implementation Details}
\label{append:impl}

We implemented the LMM in PyTorch and trained it on four GeForce RTX 2080 Ti GPUs on three commonly used benchmark datasets: CIFAR-10, ImageNet \rsm, and AFHQ \rsm (aka. AFHQ-v2 \rsm). We employed the network architectures and hyper-parameters listed in Table~\ref{tab:arch}, which were previously used in~\citep{Tero2022EDM,Song23CM,Kim2024CTM,Lipman2023flow} over these datasets.

\bf{Training Losses. } \normalfont As noted above, we used the VGG-based perceptual loss in~\citep{Johnson2016Ploss} as our image reconstruction loss term in Eq.~\ref{eq:ours}, and resized the images to 224-by-224 pixels before evaluating it. 

We use a similar adversarial loss as the one used in~\citep{Kim2024CTM}, namely, we adopted the discriminator architecture from~\citep{Sauer2022StyleGAN} and used its conditional version when training on labeled datasets. We used the same feature extraction networks they use, as well as the adaptive weighing from~\citep{Esser2O21VQGAN}, given by 
$\lambda_{\text{adapt}}  = \| \nabla_{\prm_L} \mathcal{L}_\text{lines} \| / \| \nabla_{\prm_L}  \mathcal{L}_\text{disc}\|$, where $\prm_L$ denotes the weights of the last layer of $\net_\prm$. We also used their augmentation strategy, taken from~\citep{zhao2020diffaugment}, we resized the images to 224-by-224 pixels before applying this loss as well.

\bf{Denoising ODE. } \normalfont As noted in Section~\ref{sec:method}, we use the EDM denoising score-matching model $\net^*$ in~\citep{Tero2022EDM} in order to produce our training pairs $\xs,\net^*_\smp(\xs)$. We use their deterministic sampler (second-order Heun) in order to establish a well-defined change-of-variable, $\net^*_\smp(\x)$, between the source and target distributions. This scheme uses a source distribution $\ps = \mathcal{N}(0,\sigma_{\max})$ and noise scheduling $\sigma_t = \big(\sigma_{\max}^{1/\rho}+t/(N-1)(\sigma_{\min}^{1/\rho}-\sigma_{\max}^{1/\rho})\big)^{\rho}$, where $\rho=7$ and $\sigma_{\min}=0.002$ which corresponds to a negligible noise level when reaching the target distribution, $\pt$, assuming $\Vr[\xt]$ of order around 1. This method uses $N$=18 (NFE=35) steps to draw samples from the CIFAR-10 dataset, and $N$=40 (NFE=79) for ImageNet \rsm~and AFHQ \rsm.

\bf{Sampling the LMM. } \normalfont We use the sampling scheme used in~\citep{Song23CM,Kim2024CTM} to sample the LMM. This consists of the following iterations, $\x^{t+1} = \net_\prm(\x^t,\sigma_t) + \sigma_{t+1} \eta$, where $\x^0 \sim \ps $ and $\eta \sim \mathcal{N}(0,I)$. We report the noise scheduling we use in each step, $\sigma_t$, in terms of the ones used in~\citep{Tero2022EDM}, in Tables~\ref{tab:abl_cifar}, \ref{tab:abl_imagenet}, and~\ref{tab:abl_afhq}.

\bf{Training Cost. } \normalfont The number of iterations used for training the LMM is listed in Table~\ref{tab:arch}. The first 80k pre-training iterations were executed without the ADL as well as by evaluating the VGG-perceptual loss over 64-by-64 pixel images. This made each training iteration x6 faster than the following full-resolution and using the ADL. These numbers are lower than the ones reported in~\citep{Song23CM}, 800k for CIFAR-10 and 2400k for ImageNet \rsm, and in~\citep{Kim2024CTM}, 100k for CIFAR-10 and 120k for ImageNet \rsm. We note that these methods rely on having a pre-training DSM as in our case. Training the CFM~\citep{Lipman2023flow} does not require a pre-existing model, and uses 195k iterations for CIFAR-10 and 628k for ImageNet \rsm. The numbers of iterations quoted here are normalized to a batch size of 512.

Unlike the rest of these methods, the training data of the LMM must be first generated. As noted above, it consists of pairs of the form $\xs,\net^*_\smp(\xs)$ which are sampled from the EDM model $\net^*$, in~\citep{Tero2022EDM}. The number of training examples we use for each dataset are listed in Table~\ref{tab:arch}. On one hand this sampling process uses fairly high NFEs (35 for CIFAR-10, and 79 for ImageNet \rsm~and AFHQ \rsm), but on the other hand it consists of feed-forward executions with no back-propagation calculations. Moreover, this process can be executed on single GPUs and be trivially parallelized across multiple machines. In terms of wall-clock time this pre-processing did not take long, namely, half a day for CIFAR-10 compared to the 4 days of LMM training, and six days for ImageNet \rsm~compared to 20 days of training, and two days for AFHQ \rsm~compared to 6 training days. We remind that these training sessions were conducted on four GeForce RTX 2080 Ti GPUs.


\setlength{\tabcolsep}{10pt}
\renewcommand{\arraystretch}{1} 

\begin{table}[h]
\centering
\footnotesize
\setlength{\tabcolsep}{8pt} 
\caption{Network architectures and hyper-parameters used for different datasets.}
\label{tab:arch}
\begin{tabular}{p{5cm}ccc}
\toprule
\textbf{Hyper-Parameter}  & \textbf{CIFAR-10}  &  \textbf{AFHQ \rsm} & \textbf{ImageNet \rsm} \\
\midrule
Generator architecture & DDPM++ & DDPM++ & ADM \\
Channels & 128 & 128 & 192 \\
Channels multipliers & 2, 2, 2 & 1, 2, 2, 2 & 1, 2, 3, 4 \\
Residual blocks  & 4 & 4 & 3 \\
Attention resolutions & 16 & 16 & 32, 16, 8 \\
Attention heads & 1 & 1 & 6, 9, 12 \\
Attention blocks in encoder & 4 & 4 & 9 \\
Attention blocks in decoder & 2 & 2 & 13 \\
Generator optimizer & RAdam & RAdam & RAdam \\
Discriminator optimizer & RAdam & RAdam & RAdam \\
Generator learning rate & 0.0004 & 0.0001 & 0.000008 \\
Discriminator learning rate & 0.002 & 0.002 & 0.002 \\
Generator $\beta_1, \beta_2$ & 0.9, 0.999 & 0.9, 0.999 & 0.9, 0.999 \\
Discriminator $\beta_1, \beta_2$ & 0.5, 0.9 & 0.5, 0.9 & 0.5, 0.9 \\
Batch size & 512 & 512 & 512 \\
EMA & 0.999 & 0.999 & 0.999 \\
Training images & 1M & 2M & 4M \\
Training iterations & 80k+20k w/ADL. & 80k+25k w/ADL & 80k+30k w/ADL \\
$\lambda_{\text{lines}}$ & 0.5 & 0.5 &  0.5 \\
\bottomrule
\end{tabular}
\end{table}

\begin{table}
\footnotesize
    \centering
    \setlength{\tabcolsep}{10pt} 
    \begin{tabular}{llcccc}
     \toprule
     & & \multicolumn{4}{c}{\textbf{CIFAR-10} (conditional)} \\
     \cmidrule(lr){3-6}
     & & \textbf{L2} &  \textbf{VGG} & \textbf{VGG+ADL} & \textbf{VGG+ADL+SOT}  \\
    \textbf{NFE} & \textbf{Steps} & \textbf{FID} $\boldsymbol{\pm}$ \textbf{std} & \textbf{FID} $\boldsymbol{\pm}$ \textbf{std} & \textbf{FID} $\boldsymbol{\pm}$ \textbf{std} & \textbf{FID} $\boldsymbol{\pm}$ \textbf{std} \\
    \midrule
    1 & 0 & 5.125 $\pm$ 0.050 & 3.124 $\pm$ 0.024 & 1.672 $\pm$ 0.018 & 1.575  $\pm$ 0.016 \\
    2 & 0, 1 & 4.289 $\pm$ 0.032 & 2.796 $\pm$ 0.020 & 1.394 $\pm$ 0.010 & 1.389 $\pm$ 0.011 \\
    3 & 0, 1, 2 & 4.019 $\pm$ 0.026 & 2.761 $\pm$ 0.021 & 1.386 $\pm$ 0.009 & - \\
    3 & 0, 3, 5 & 3.337 $\pm$ 0.042 & 2.601 $\pm$ 0.019 & 1.381 $\pm$ 0.015 & - \\
    4 & 0, 1, 3, 5 & 3.315 $\pm$ 0.023 & 2.625 $\pm$ 0.025 & 1.383 $\pm$ 0.012 & - \\
    \bottomrule
    \end{tabular}
    \caption{Selected step indices $t$ from the original EDM schedule $\sigma_t$ consisting of 18 steps for this dataset.}
    \label{tab:abl_cifar}
\end{table}

\begin{table}
\footnotesize
    \centering
    \begin{tabular}{llccc}
    \toprule
    & &  \multicolumn{3}{c}{\textbf{ImageNet \rsm~} (conditional)} \\
    \cmidrule(lr){3-5} 
     & &  \textbf{VGG} & \textbf{VGG+ADL} & \textbf{VGG+ADL+SOT}  \\
    \textbf{NFE} & \textbf{Steps} &   \textbf{FID} $\boldsymbol{\pm}$ \textbf{std} &   \textbf{FID} $\boldsymbol{\pm}$ \textbf{std} &   \textbf{FID} $\boldsymbol{\pm}$ \textbf{std} \\
    \midrule
    1 & 0 & 6.968 $\pm$ 0.051 & 1.731 $\pm$ 0.013 &  1.473 $\pm$ 0.016 \\
    2 & 0, 1 & 5.472 $\pm$ 0.042 & 1.318 $\pm$ 0.013 & 1.167 $\pm$ 0.016 \\
    3 & 0, 1, 2 & 5.004 $\pm$ 0.057 & 1.301 $\pm$ 0.012 & - \\
    3 & 0, 3, 5 & 4.694 $\pm$ 0.047 & 1.284 $\pm$ 0.016 & -\\
    \bottomrule
    \end{tabular}
    \caption{Selected step indices $t$ from the original EDM schedule $\sigma_t$ consisting of 40 steps for this dataset.}
    \label{tab:abl_imagenet}
\end{table}

\begin{table}
\footnotesize
    \centering
    \begin{tabular}{llccc}
    \toprule
    & &  \multicolumn{3}{c}{\textbf{AFHQ \rsm}} \\
    \cmidrule(lr){3-5} 
     & &  \textbf{VGG} & \textbf{VGG+ADL} & \textbf{VGG+ADL+SOT}  \\
    \textbf{NFE} & \textbf{Steps} &   \textbf{FID} $\boldsymbol{\pm}$ \textbf{std} &   \textbf{FID} $\boldsymbol{\pm}$ \textbf{std} &   \textbf{FID} $\boldsymbol{\pm}$ \textbf{std} \\
    \midrule
    1 & 0 &  5.458 $\pm$ 0.053 & 2.687 $\pm$ 0.046 & 2.767 $\pm$ 0.056 \\
    2 & 0, 1 &  4.254 $\pm$ 0.039 & 1.545 $\pm$ 0.023 & 1.776 $\pm$ 0.022 \\
    3 & 0, 1, 2 & 4.165 $\pm$ 0.045 & 1.462 $\pm$ 0.016& - \\
    3 & 0, 3, 5 &  3.919 $\pm$ 0.035 & 1.447 $\pm$ 0.023 & - \\
    \bottomrule
    \end{tabular}
    \caption{Selected step indices $t$ from the original EDM schedule $\sigma_t$ consisting of 40 steps for this dataset.}
    \label{tab:abl_afhq}
\end{table}

\begin{table}[]
    \centering
    \footnotesize
    
\begin{tabular}{c}
    \setlength{\tabcolsep}{3pt}
    \begin{tabular}{cccccccccc}
         \multirow{2}{*}{\textbf{EDM}} & \multicolumn{4}{c}{\textbf{LMM}} &  \multirow{2}{*}{\textbf{EDM}} & \multicolumn{4}{c}{\textbf{LMM}} \\\cmidrule(lr){2-5} \cmidrule(lr){7-10} 
            & \multicolumn{2}{c}{ wo/ADL} & \multicolumn{2}{c}{ w/ADL} & & \multicolumn{2}{c}{ wo/ADL} & \multicolumn{2}{c}{ w/ADL} \\ \cmidrule(lr){2-3}\cmidrule(lr){4-5} \cmidrule(lr){7-8}\cmidrule(lr){9-10}
         NFE 35 & NFE 1 & NFE 2 & NFE 1 & NFE 2 &  NFE 79 & NFE 1 & NFE 2 & NFE 1 & NFE 2 \\
    \end{tabular} \vspace{0.02in} \\
    \includegraphics[width=4.3in]{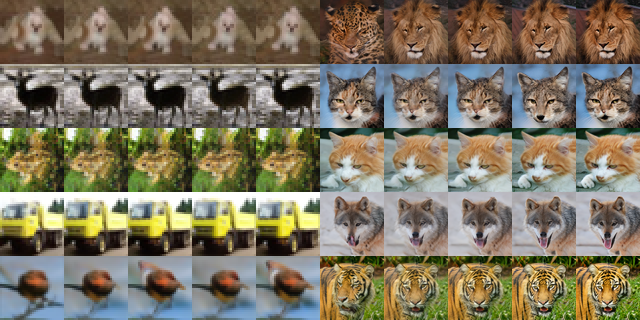}
    \end{tabular}
    \caption{CIFAR-10 (left) and AFHQ \rsm~(right) Samples Comparison.}
    \label{tab:comp2}
\end{table}

\end{document}